\def\year{2020}\relax
\documentclass[letterpaper]{article} 
\usepackage{aaai20}  
\usepackage{times}  
\usepackage{helvet} 
\usepackage{courier}  
\usepackage[hyphens]{url}  
\usepackage{graphicx} 
\usepackage{graphicx}  
\usepackage{subcaption}
\usepackage{multirow}
\usepackage{amsmath}
\usepackage{array}
\title{FD-GAN: Generative Adversarial Networks with Fusion-discriminator \\for Single Image Dehazing}
\author{
Yu Dong\textsuperscript{\rm 1 2}\thanks{The first two authors contributed equally and should be regarded as co-first authors.} \space \space 
Yihao Liu\textsuperscript{\rm 1 2}\footnotemark[1]  \space \space 
He Zhang\textsuperscript{\rm 3} \space \space 
Shifeng Chen\textsuperscript{\rm 1}\thanks{Corresponding author.}   \space \space 
Yu Qiao\textsuperscript{\rm 1}\\ 
\textsuperscript{\rm 1}ShenZhen Key Lab of Computer Vision and Pattern Recognition,\\ Shenzhen Institutes of Advanced Technology, Chinese Academy of Sciences\\
\textsuperscript{\rm 2}University of Chinese Academy of Sciences \space\space\space
\textsuperscript{\rm 3}Adobe Inc.\\ 
\{yu.dong, shifeng.chen, yu.qiao\}@siat.ac.cn, \\
liuyihao14@mails.ucas.ac.cn, hezhan@Adobe.com 
}
 
\begin{document}

\maketitle

\begin{abstract}
Recently, convolutional neural networks (CNNs) have achieved great improvements in single image dehazing and attained much attention in research. Most existing learning-based dehazing methods are not fully end-to-end, which still follow the traditional dehazing procedure: first estimate the medium transmission and the atmospheric light, then recover the haze-free image based on the atmospheric scattering model. However, in practice, due to lack of priors and constraints, it is hard to precisely estimate these intermediate parameters. Inaccurate estimation further degrades the performance of dehazing, resulting in artifacts, color distortion and insufficient haze removal. To address this, we propose a fully end-to-end Generative Adversarial Networks with Fusion-discriminator (FD-GAN) for image dehazing. With the proposed Fusion-discriminator which takes frequency information as additional priors, our model can generator more natural and realistic dehazed images with less color distortion and fewer artifacts. Moreover, we synthesize a large-scale training dataset including various indoor and outdoor hazy images to boost the performance and we reveal that for learning-based dehazing methods, the performance is strictly influenced by the training data. Experiments have shown that our method reaches state-of-the-art performance on both public synthetic datasets and real-world images with more visually pleasing dehazed results.
\end{abstract}

\section{Introduction}
Haze is a common atmospheric phenomenon where dust, smoke or other floating particles greatly absorb and scatter the light, resulting in degradations in image quality. Hazy images usually lose contrast, color fidelity and edge information, which can reduce the visibility of scenes and further do harm to plenty of computer vision tasks and relevant applications such as classification, localization and self-driving system. Hence, haze removal is highly desired. There is a classical atmospheric scattering model \cite{8} widely used to describe the formation of a hazy image:
\begin{equation}
I(x)=J(x)t(x)+A(1-t(x)),
\end{equation}
where $I(x)$ is the observed hazy image, $J(x)$ is the real scene radiance to be recovered, $t(x)$ is the medium transmission map and $x$ indicates the pixel coordinate. $A$ is the global atmospheric light. If the global atmosphere is homogenous, the transmission map $t(x)$ can be modeled by \begin{equation}
t(x)=e^{-\beta d(x)},
\end{equation}
where $d(x)$ represents the scene depth and $\beta$ is the scattering coefficient of the atmosphere. As only $I(x)$ is given, recovering the haze-free scene $J(x)$ is an ill-posed problem.

\begin{figure}[t]
	\centering
	\includegraphics[scale=0.305]{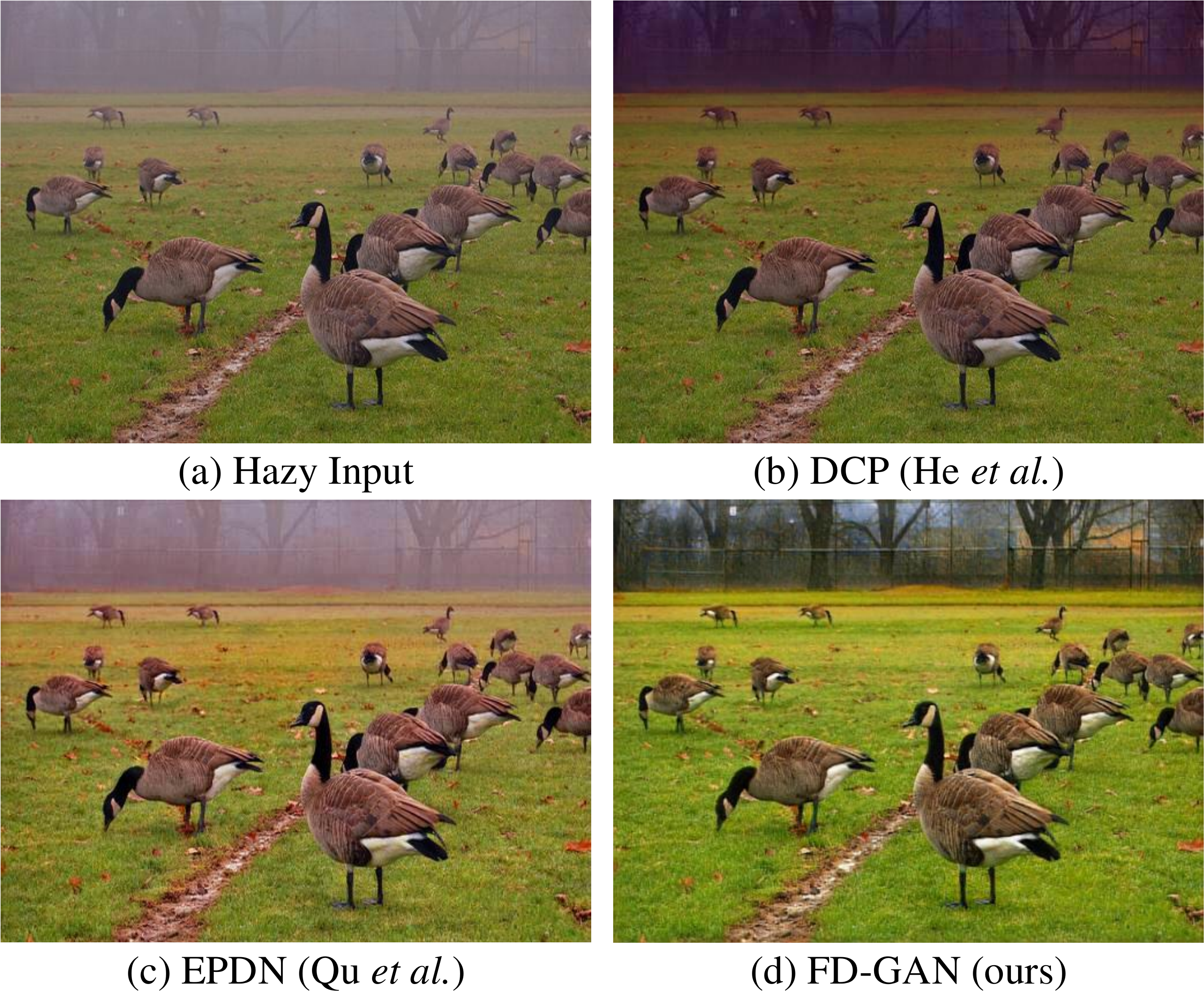}
	\caption{Dehazing example on a real-world hazy image. Comparing with existing methods, our model can remove haze effect more thoroughly and generate more natural and visually pleasing dehazed results with less color distortion.}
\end{figure}

In recent years, numerous methods have been proposed to tackle this challenging problem. These methods can be divided into two trends: prior-based methods and learning-based methods. In prior-based methods, the global atmospheric light $A$ and the medium transmission map $t(x)$ are evaluated by hand-crafted priors. These priors are supposed to distinguish between hazy images and haze-free images, such as dark channel prior \cite{3} and color attenuation prior \cite{32}. Learning-based methods utilize convolutional neural networks (CNNs) to extract features and learn the mappings between hazy and haze-free image pairs from enormous training data. At present, most of learning-based methods still follow the conventional procedure of dehazing: first estimate the medium transmission $t(x)$ and the atmospheric light $A$, then recover the haze-free image based on the atmospheric scattering model. However, due to lack of real-world training data and effective priors, it is difficult to precisely estimate these intermediate parameters. Inaccurate estimation will further degrade the performance of dehazing. 

Accordingly, we propose a fully end-to-end Generative Adversarial Networks with Fusion-discriminator (FD-GAN) for image dehazing, which takes the hazy image as input and directly generates the haze-free image without the estimation of intermediate parameters. It leverages the power of generative adversarial network (GAN) and makes effective use of prior knowledge by the proposed Fusion-discriminator that integrates the frequency information into the design of the network.

As an image can be decomposed into high-frequency (HF) and low-frequency (LF) components, we integrate them as additional priors and constraints into the discriminator. Specifically, we concatenate the generated image $G(I)$ (or ground truth image $J$) and its corresponding LF and HF as a sample, then feed it into the discriminator where we name it Fusion-discriminator. Different from traditional discriminator that only learns to distinguish between the generated image and the ground truth image, the proposed Fusion-discriminator additionally takes LF and HF information into account. Our method can move haze effect more thoroughly and generate dehazed results with less color distortion. Figure 1 shows a dehazing example of our proposed method.

For data-driven approaches, training data is crucial to the final performance. However, most of the dehazing data are synthesized from NYU dataset \cite{61}, which only includes indoor images. Although several outdoor datasets have been adopted recently, the scenes diversity and quantity of them are limited, such as HazeRD \cite{89}, Make3D \cite{88}. To address this, we synthesize a new training dataset based on Microsoft COCO dataset \cite{27} including indoor and outdoor images. Experiments show the model trained on this dataset performs much better than that trained on other commonly used datasets.

In summary, our contributions are as follows:
\begin{quote}
	\begin{itemize}
		\item We propose a fully end-to-end image dehazing algorithm FD-GAN, which can directly output haze-free images without the estimation of intermediate parameters. It achieves state-of-the-art performance on both synthetic and real-world scenes.
		\item We develop a novel Fusion-discriminator which integrates the frequency information as additional priors and constraints into the learning procedure. Experiments demonstrate that the Fusion-discriminator has great capability of facilitating the generator to produce more realistic and clearer dehazed images.
		\item We demonstrate that the performance of learning-based dehazing algorithms is critically affected by the quality of training data. Hence, we propose a new synthetic dataset for training, which includes various indoor and outdoor scenes. Based on this dataset, our model could be further improved and surpass other existing methods qualitatively and quantitively.
	\end{itemize}
\end{quote}

\begin{figure}[h]
	\centering
	\includegraphics[scale=0.2]{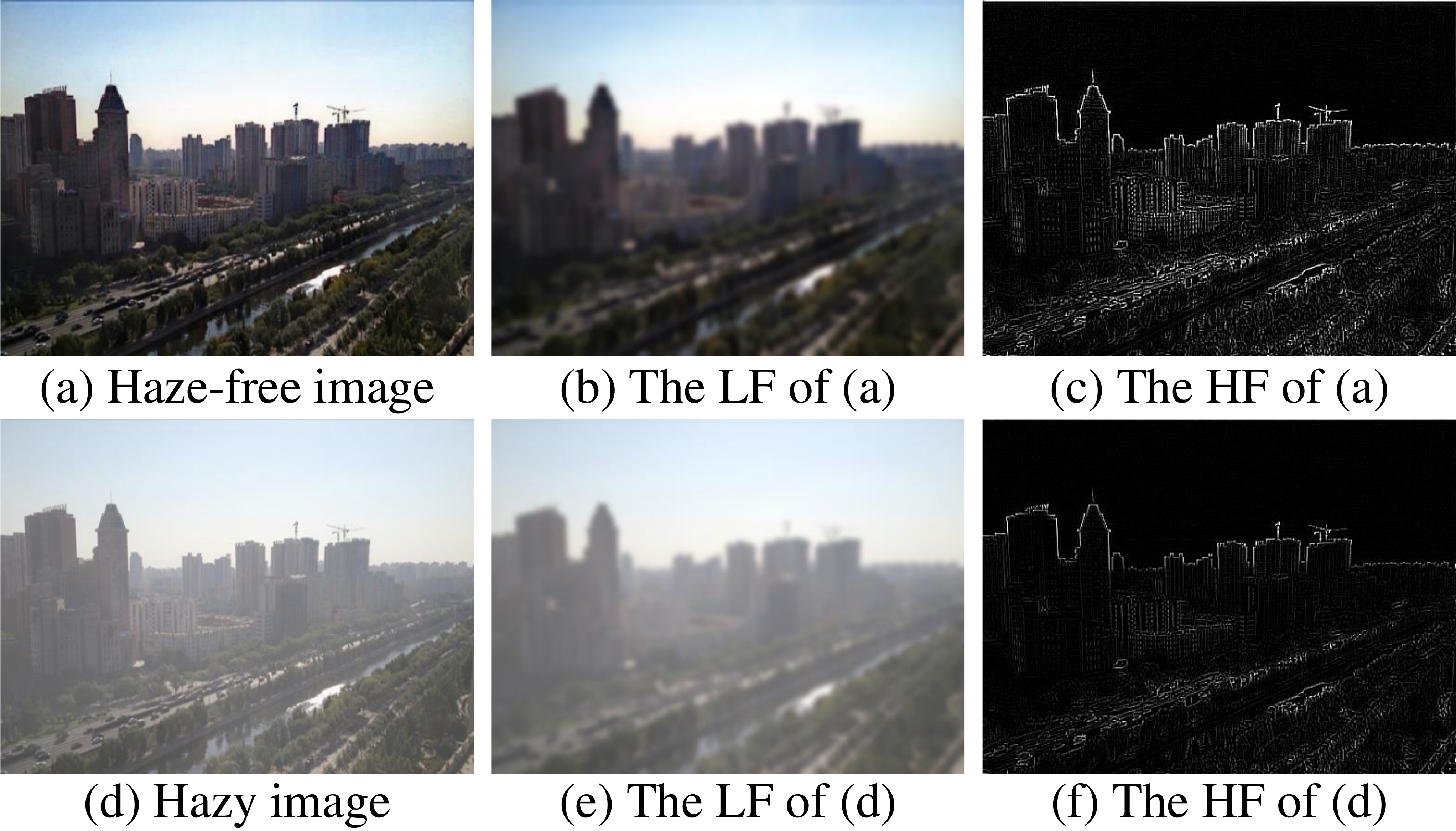}
	\caption{Hazy images usually lose contrast, color saturation and lack edge information. It can be observed that the HF and LF of hazy and haze-free images differ from each other.}
\end{figure}

\section{Related Work}
\textbf{Image dehazing.} In recent decades, numerous dehazing methods have been proposed. These methods can be roughly divided into two categories: prior-based methods and learning-based methods.
\textit{Prior-based methods} use hand-designed priors or assumptions for image dehazing. Based on the observation that clear images usually have higher contrast, Tan \cite{6} proposed a patch-based contrast maximization method. Fattal \cite{1} estimated the albedo of the scene, for thinking that the transmission and surface shading are locally uncorrelated. He \textit{et al.} \cite{3} proposed a classical statistical observation that in majority of non-sky regions, at least one color channel has extremely low intensities in some pixels. Based on this prior, the transmission map and atmospheric light can be estimated.
\textit{Learning-based methods} utilize convolution neural networks (CNNs) to extract features  for dehazing from enormous training data. Cai \textit{et al.} \cite{9} introduced CNN for estimating the transmission map and atmospheric light. Li \textit{et al.} \cite{41} proposed a novel network to learn the transmission map and atmospheric light simultaneously. Zhang \textit{et al.} \cite{16} built two CNN branches for estimating the transmission map and atmospheric light, respectively. Suárez \textit{et al.} \cite{90} explored fully end-to-end dehazing network using cGAN. Qu \textit{et al.} \cite{75} treated the dehazing task as an image-to-image translation problem. They embedded GAN with an enhancer to reinforce the dehazing effect. Nevertheless, as dehazing is a highly ill-posed and under-constrained problem, these methods have not taken effective priors into account, leading the reconstructed images likely to contain color distortion, artifacts and insufficient haze removal.\\
\textbf{Generative Adversarial Networks.} Generative Adversarial Networks (GANs) \cite{46} consist of two parts: the generator and the discriminator, contesting with each other in a zero-sum game framework. With continuous research, cGAN \cite{65}, DCGAN \cite{66} and WGAN \cite{67} were proposed to boost the performance and maintain the stability of the training process. GAN has great potential to generate realistic images and have been introduced in various vision tasks, such as super-resolution \cite{80}, de-raining \cite{91}.

\begin{figure*}[t]
	\centering
	\begin{subfigure}{0.8\linewidth}
		\centerline{\includegraphics[width=1.0\columnwidth]{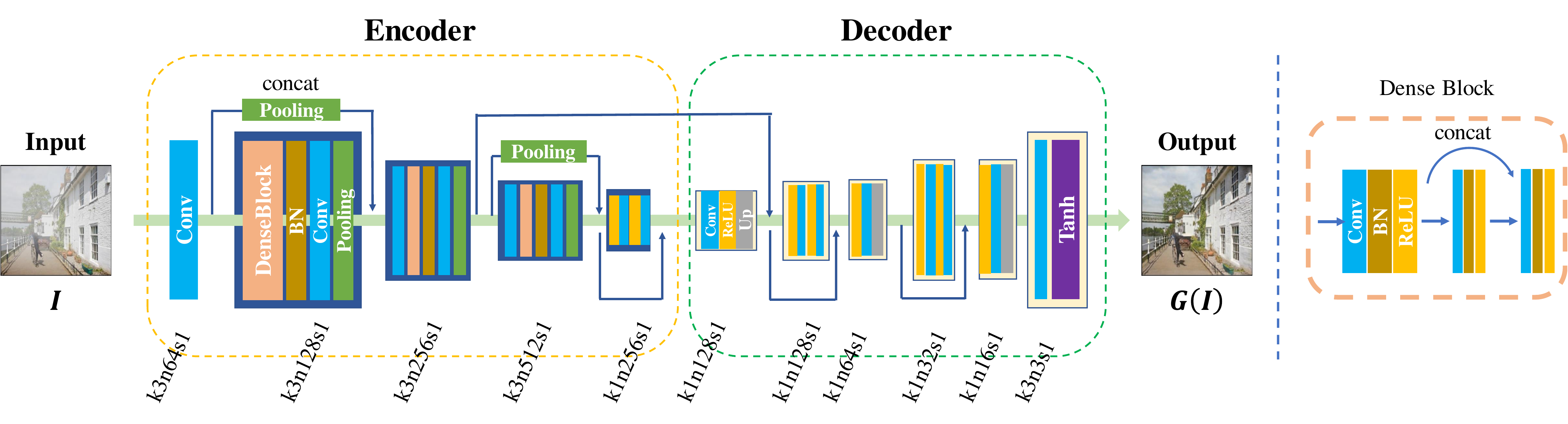}}
		\subcaption{Generator. The generator can directly output the haze-free images without estimation of intermediate parameters. The densely connected encoder-decoder can maximize the information flow between layers in the network.}
	\end{subfigure}
	
	\begin{subfigure}{0.8\linewidth}
		\centerline{\includegraphics[width=1.0\columnwidth]{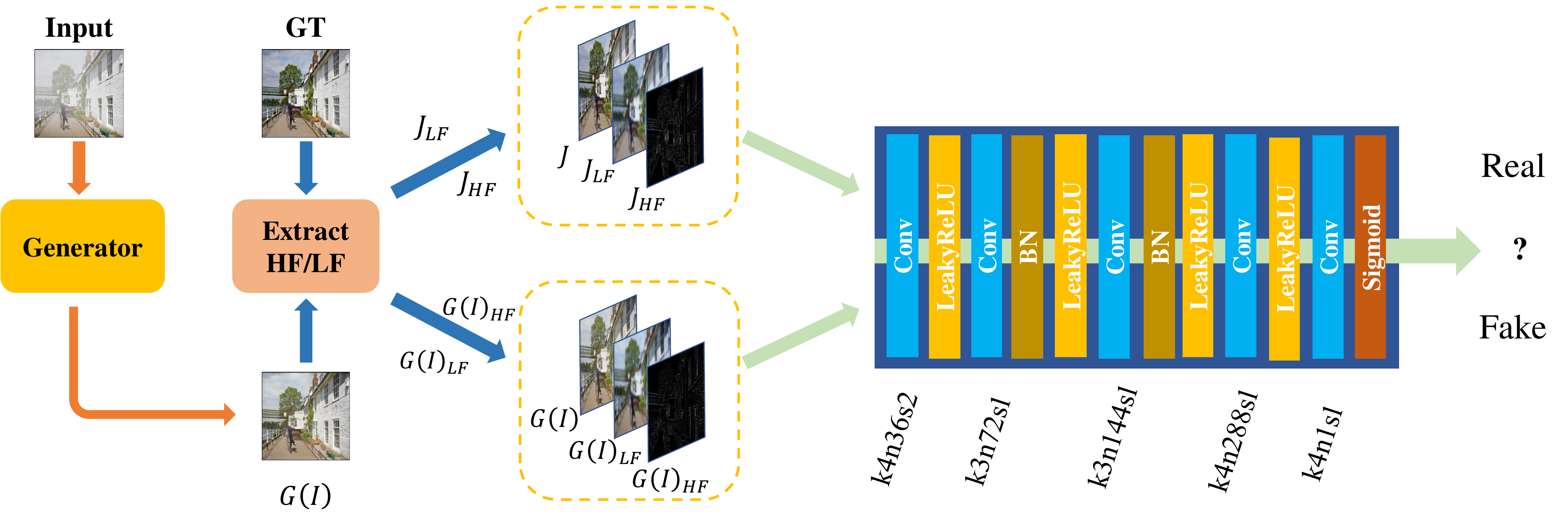}}
		\subcaption{Fusion-discriminator. The frequency information is integrated as additional prior and supervision into the discriminator, which can guide the generator to generate more realistic dehazed results with less color distortion.}
	\end{subfigure}	
	\caption{Architecture of our proposed method.}
	\label{figure3}
\end{figure*}

\section{Method}
In this section, we introduce our method in detail including the generator, the Fusion-discriminator and studies on the variants of the Fusion-discriminator.

\subsection{Densely Connected Encoder-decoder}
In our design, the aim of the generator is to directly generate clear images from hazy inputs, without the stage of estimating intermediate parameters. To achieve this goal, the generator is supposed to preserve the image content and recover the details as much as possible, while removing the haze. Several works have demonstrated dense connections have the potential to facilitate the extraction and utilization of features, especially for low-level vision tasks \cite{82}. Inspired by these previous works, we design a densely connected encoder-decoder as the generator, which can make full use of all the features extracted from shallow layers to deep layers. As shown in Figure 3(a), the encoder contains three dense blocks, in which a series of convolutional, batch normalization (BN) and ReLU layers are included. After pooling operation, the feature map size (height and width) gradually shrinks to 1/8 of the input size. In the decoder module, the size of feature maps is gradually recovered to the original resolution. We choose nearest-neighbor interpolation for up-sampling, since it is reported to be less likely to produce checkerboard artifacts \cite{73}.

\begin{figure*}[!htbp]
	\centering
	
	\includegraphics[width=1.8\columnwidth]{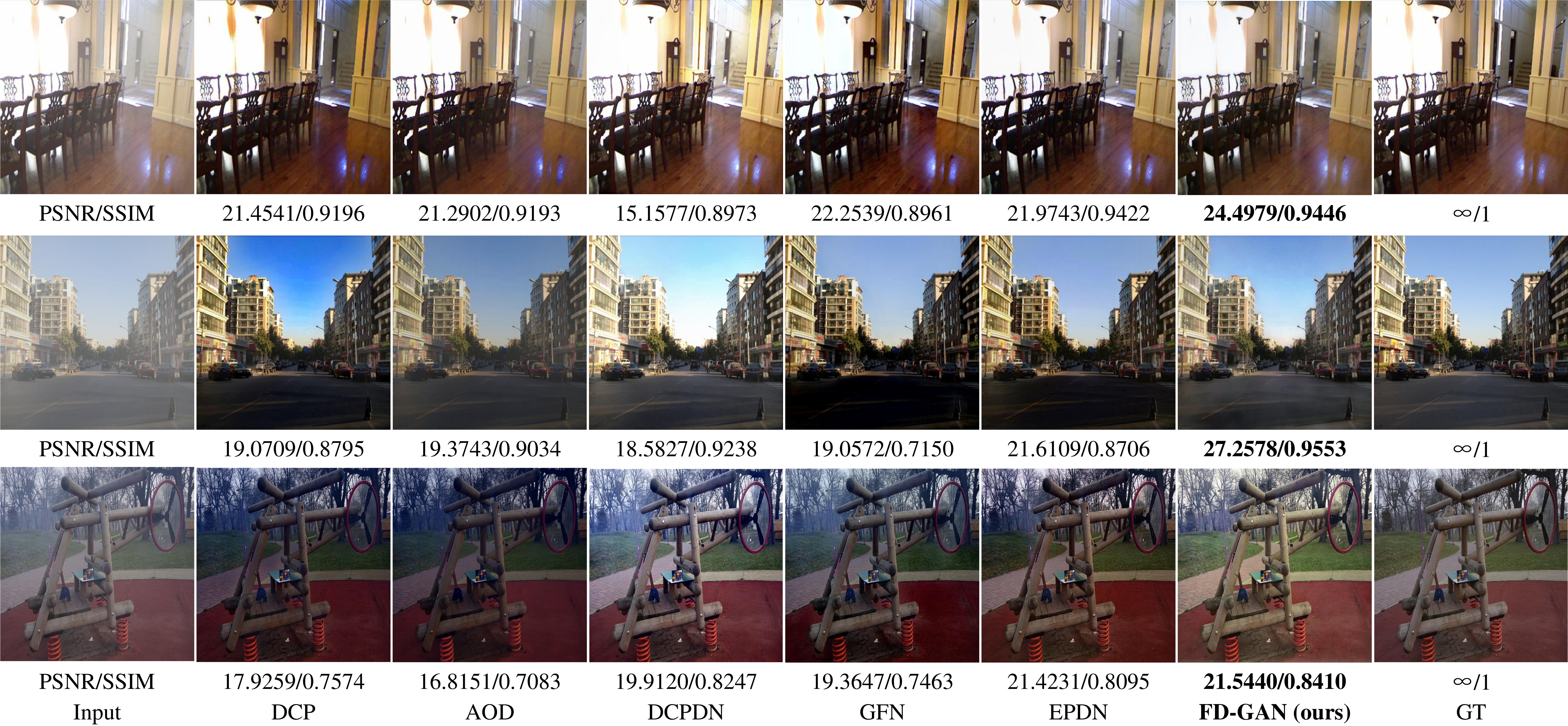}	
	\centering
	\caption{Visual comparisons on synthetic datasets. Our method yields highest PSNR and SSIM on public synthetic datasets.}
\end{figure*}

\subsection{Fusion-discriminator}
For hazy images and clear images, there are many discrepancies between them. For example, haze-free images usually have higher contrast and sharper edges compared with hazy images. Lots of dehazing methods lies on such priors or assumptions. However, these priors do not hold good for all conditions and can be easily violated in practice. In our method, we integrate the frequency information as basic priors and additional constraints into the design of the discriminator.

In computer vision and signal processing, an image can be decomposed into high-frequency (HF) and low-frequency (LF) components. HF and LF are basic characteristics of images, in which hazy images and clear images differ from each other. HF component represents those image parts whose intensity changes rapidly, such as sharp edges, textures and fine details. On the contrary, image parts whose intensity values change slowly are LF regions, i.e., smooth regions. With the high-frequency details removed, LF stresses the brightness, color and contrast information of images, and it can make color comparison easier \cite{72}. We adopt this idea and develop a novel Fusion-discriminator that takes both HF and LF information as additional priors and constraints for dehazing. The HF and LF can assist the discriminator to distinguish the differences in textures and major color between hazy and haze-free images. To extract the LF information, we apply a Gaussian filter on the image to remove the high-frequency details. For HF component, we apply a Laplace operator on the image, after which the edges and textures will be stressed. Figure 2 gives an example of the extracted HF and LF of a hazy image and a dehazed image.

As depicted in Figure 3(b), given an input hazy image $I$ and the ground truth image $J$, the output of generator is $G(I)$. We then concatenate $G(I)$ with its corresponding LF component $G(I)_{LF}$ and HF component $G(I)_{HF}$ as a sample $[G(I), G(I)_{LF}, G(I)_{HF}]$. When training the discriminator, the concatenated sample $[G(I), G(I)_{LF}, G(I)_{HF}]$ is labeled as \textit{fake}, while $[J, J_{LF}, J_{HF}]$ is labeled as \textit{real}. Thus, our method can be formulated as the following min-max optimization problem:
\begin{equation}
\begin{aligned}
&\mathop {\min }\limits_G \mathop{\max}\limits_{D} E_{J\sim p_{train}(J)} \big[\log{D}(J \bowtie J_{LF}\bowtie J_{HF})\big]\\
&+E_{I \sim p_{G} (I)} \big[\log(1-D(G(I) \bowtie G(I)_{LF} \bowtie G(I)_{HF}))\big],
\end{aligned}
\end{equation}
where $\bowtie$ refers to the concatenation operation. By fusing the low-frequency and high-frequency prior into the discriminator network, our method can generate more natural and realistic dehazed images with less color distortion and fewer artifacts, as shown in Figure 4 and Figure 5.

\begin{figure*}[htbp]
	\centering
	\includegraphics[width=2.1\columnwidth,height=0.55\linewidth]{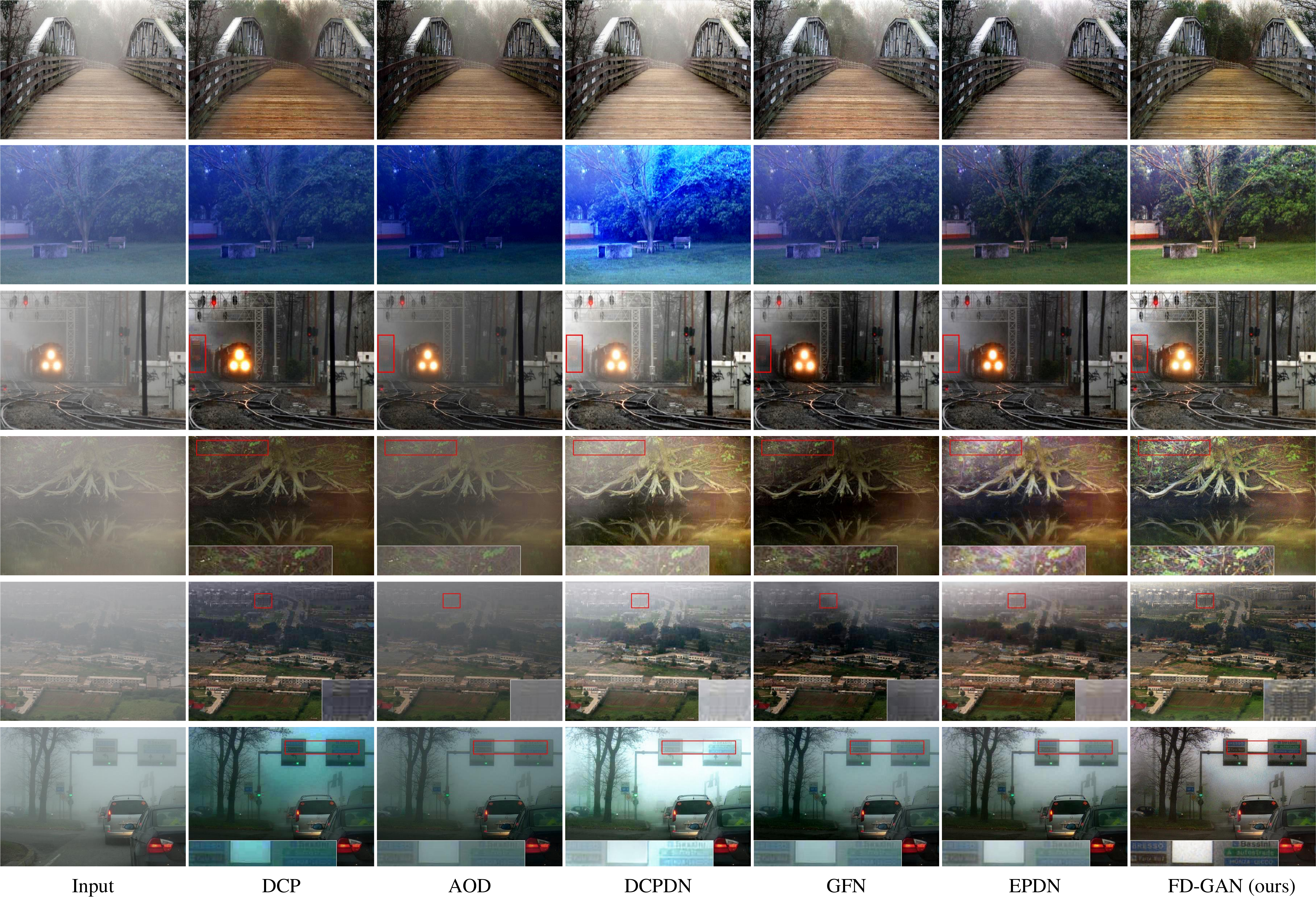}	
	
	\centering
	\caption{Visual comparisons on real-world hazy images. Our model can generate more natural and visual pleasing dehazed results with less color distortion. Please see the details in red rectangles. Zoom in for best view.}
\end{figure*}

\begin{figure}[!h]
	\centering
	\includegraphics[width=1\columnwidth]{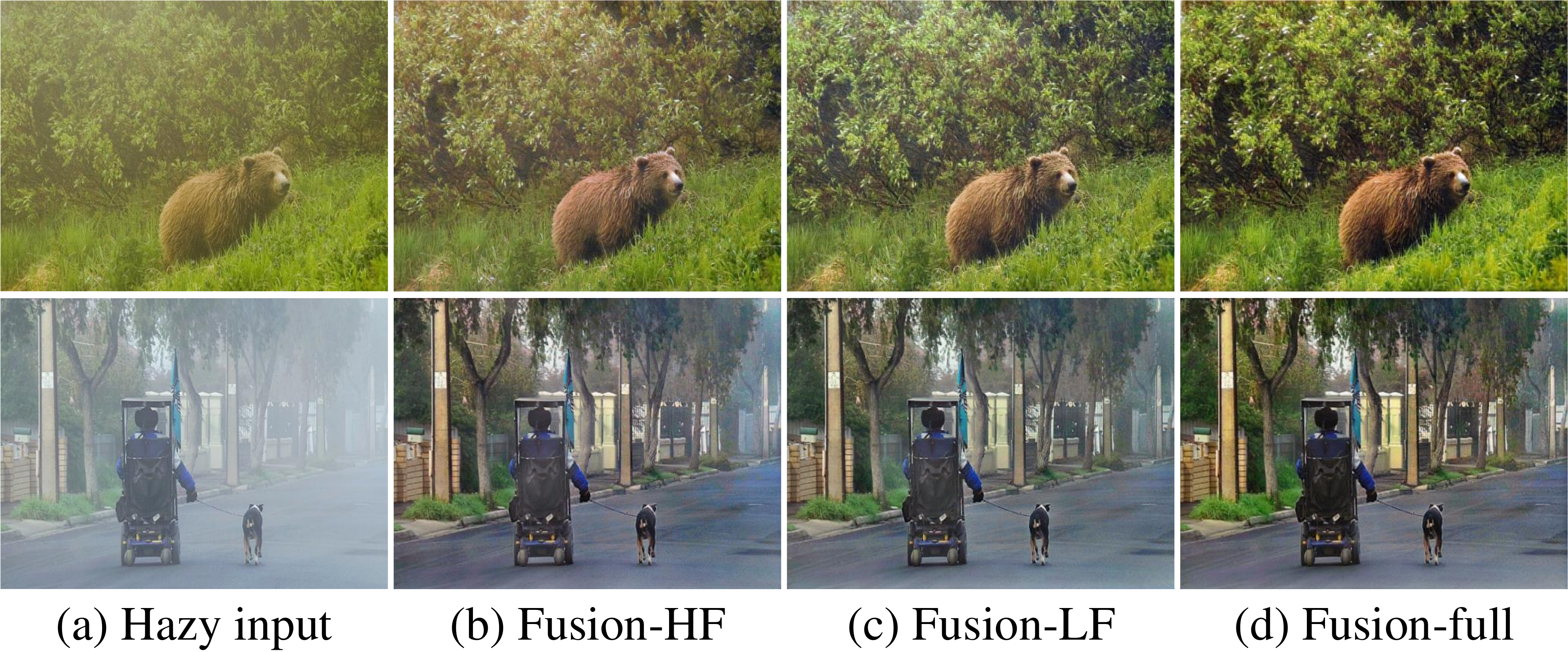}	
	\caption{Visual results of different variants of Fusion-discriminator. Fusion-full model can successfully combine the strengths of other models.}
	\label{figure2}
\end{figure}

\linespread{1.2}
\begin{table}[h]
	\centering
	\caption{Quantitative comparisons with different variants of Fusion-discriminator.}\smallskip
	
	\begin{tabular}{p{50pt}||p{68pt}<{\centering}|p{68pt}<{\centering}}
		\hline
		\small \multirow{2}{*}{Method} & \small SOTS & \small NTIRE'18 \\ 
		& \small \underline{PSNR}/\underline{SSIM} & \small \underline{PSNR}/\underline{SSIM} \\ \hline
		Fusion-HF & 21.7989/0.9035 & 16.8903/0.7108 \\
		Fusion-LF & 22.1635/0.9150 & 17.6703/0.7211 \\
		Fusion-full & \textbf{23.1529}/\textbf{0.9207} &  \textbf{18.0684}/\textbf{0.7300} \\
		\hline
	\end{tabular}
	\label{tab:table1}
\end{table}

\linespread{1.2}
\begin{table}[h]
	\centering
	\caption{Quantitative comparisons with other state-of-the-art methods on public synthetic testing datasets.}\smallskip
	
	\begin{tabular}{p{81pt}||p{60pt}<{\centering}|p{60pt}<{\centering}}
		\hline
		\small \multirow{2}{*}{Method} & \small SOTS & \small NTIRE'18 \\ 
		& \small \underline{PSNR}/\underline{SSIM} & \small \underline{PSNR}/\underline{SSIM} \\ \hline
		\small DCP (He \textit{et al.})  & \small{18.3419/0.8678} & \small{14.2143/0.6243} \\
		\small AOD (Li \textit{et al.}) & \small{19.7840/0.8668} & \small{15.2504/0.6122} \\
		\small DCPDN (Zhang \textit{et al.}) & \small{17.9777/0.8565} & \small{13.8651/0.6856} \\
		\small GFN (Ren \textit{et al.})  & \small{21.8147/0.8614} & \small{15.0671/0.6191} \\
		\small EPDN (Qu \textit{et al.}) & \small{22.6783/0.9015} & \small{16.6053/0.6926} \\
		\small \textbf{FD-GAN (ITS)} & \small{22.8069/0.8889} & \small{16.6008/0.6953} \\
		\small \textbf{FD-GAN (OTS)} & \small{22.8212/0.8886} & \small{16.3797/0.7146} \\
		\small \textbf{FD-GAN (ours)} & \small{\textbf{23.1529}/\textbf{0.9207}} &  \small{\textbf{18.0684}/\textbf{0.7300}} \\
		\hline
	\end{tabular}
	\label{tab:table1}
\end{table}

\subsection{Discussions on Fusion-discriminator}
Traditional discriminators only learn to distinguish between generated image and ground truth image, which is a relatively weak supervision for searching optimal solution. To further improve the countervailing ability of the network, the discriminator is supposed to distinguish more features and characteristics of real and fake data, so that the generator can generate more realistic results that are closer to the ground truth.
We investigate several variants of the Fusion-discriminator to further explore its capability. Previously the input of Fusion-discriminator is $[G(I), G(I)_{LF}, G(I)_{HF}]\big/[J, J_{LF}, J_{HF}]$, which we will denote as \textit{Fusion-full} in the following. This model leverages both HF and LF information at the same time. What if we only use HF or LF as additional constraint integrated into the network? Which component is more effective for dehazing?
To answer this question, we utilize $[G(I), G(I)_{LF}]\big/[J, J_{LF}]$ or $[G(I), G(I)_{HF}]\big/[J, J_{HF}]$ as a new sample to feed into the Fusion-discriminator. Under this protocol, the model only concatenates one of the LF and HF component, which we denote as \textit{Fusion-LF} and \textit{Fusion-HF}, respectively. 
Except the input of the Fusion-discriminator, the rest of the network is kept the same. The quantitative results are shown in Table 1 and visual results are shown in Figure 6. From the results, we can draw the following observations: 1) Fusion-full model reaches the highest PSNR and SSIM values on both SOTS and NTIRE’18 datasets followed by Fusion-LF model. 2) The visual performance on real-world images of each model are comparable with each other. They all can remove the haze effect and generate visually pleasing dehazed results. 3) Compared with Fusion-LF, Fusion-HF moves more haze effect but tends to produce a little color distortion in objects. 4) Since LF stresses the major color, contrast and structure of an image, it is not surprising that the dehazed results of Fusion-LF have better color fidelity and contrast. 5) Fusion-full model can successfully combine the strengths of Fusion-HF and Fusion-LF models and generate more natural results with better contrast, luminance and color fidelity. Intuitively, HF stresses the edges and textures of images and hazy images usually lose edge information, which could help the discriminator to detect and distinguish the hazy regions. Thus, Fusion-HF could remove more haze on fine-grained objects. As for LF, it stresses the major color and contrast information of images. Thus, with LF constraint, the Fusion-LF model can generate visually pleasing results with less color distortion. It is shown that different variants can show different characteristic according to the different fusion information. Overall, all three variants of the Fusion-discriminator can obtain superior performance which embodies the effectiveness of such architecture.

\subsection{Loss Function}
\textbf{Pixel-wise loss.} Given an input hazy image $I_i$, the output of generator is $G(I_i)$ and the ground truth is $J_i$. Then the $L_{1}$ norm loss over $N$ samples can be written as:
\begin{equation} 
L_{1}=\sum_{i=1}^N\| G(I_{i})-J_{i}\|_1,
\end{equation}
It measures the distortion/fidelity between the dehazed image and the ground truth in image pixel space.\\
\textbf{SSIM loss.} SSIM is proposed to measure the structural similarity between two images. It can be written as:
\begin{equation}
SSIM(G(I),J)=\frac{2\mu_{G(I)}\mu_{J}+C_{1}}{\mu_{G(I)}^2+\mu_J^2+C_{1}} \cdot \frac{2\sigma_{G(I) J}+C_{2}}{\sigma_{G(I)}^2+\sigma_J^2+C_{2}},
\end{equation}
where $\mu_{x}$, $\sigma_{x}^{2}$ are the average value and the variance of $x$, respectively. $\sigma_{xy}$ is the covariance of $x$ and $y$. $C_{1}$, $C_{2}$ are constants used to maintain stability. SSIM ranges from 0 to 1. SSIM loss is defined as follows:
\begin{equation}
L_{S}=1-SSIM(G(I),J).
\end{equation}
\textbf{Perceptual loss.} Besides the pixel-wise loss, we adopt perceptual loss \cite{58} to measure the perceptual similarity in feature space:
\begin{equation} 
L_{p}=\sum_{i=1}^N\| \phi(G(I_{i}))-\phi(J_{i})\|_1,
\end{equation}
where $\phi(.)$ represents the feature maps obtained by the $Relu1\_2$ layers within the VGG16 network.\\
\textbf{Adversarial loss.}  Thanks to adversarial learning, GAN has great potential to generate realistic results:
\begin{equation}
L_{G} = \log(1 - D_{fusion}(G(I) \bowtie G(I)_{LF} \bowtie G(I)_{HF})),
\end{equation}
where $\bowtie$ denotes the concatenation operation. This adversarial loss enforces our network to favor solutions that reside on the manifold of natural haze-free images.\\
\textbf{Integral loss function.} Finally, we combine the pixel-wise loss, perceptual loss and adversarial loss together to regularize the dehazing network.
\begin{equation}
L =\alpha_1 L_{1}+\alpha_2 L_{S}+\alpha_3 L_{p}+\alpha_4 L_{G}
\end{equation}
where $\alpha_1$, $\alpha_2$, $\alpha_3$ and $\alpha_4$ are positive weights.

\section{Experiment}
\subsection{Dataset}
\textbf{Dataset for training.} With regard to data-driven approaches, training data is crucial to the final performance. For lots of existing learning-based methods, the training datasets are synthesized from NYU dataset \cite{61}, which only includes indoor images. To address this, we synthesize a new dehazing dataset based on Microsoft COCO dataset \cite{27}, including various indoor and outdoor hazy images. We utilize the method in MegaDepth \cite{26} to generate the depths of image scenes and adopt similar ways in MSCNN \cite{38} to synthesize the training hazy images. The atmospheric light is randomly sampled in [0.5, 1], and the scattering coefficient is randomly set within [1.2, 2.0]. Totally we obtain 24, 200 paired synthetic images.\\
\textbf{Dataset for testing.} For testing, we use two public datasets for evaluation: Synthetic Objective Testing Set (SOTS) of RESIDE \cite{86} and NTIRE’18 \cite{87}. SOTS contains 500 indoor and 500 outdoor synthetic hazy images (non-overlapping with RESIDE Training sets). NTIRE’18 was distributed by NTIRE 2018 challenge on image dehazing. Considering the input size limit for some methods, the test images of SOTS and NTIRE'18 are resized to $512 \times 512$ and $1024 \times 1024$, respectively. The Peak Signal to Noise Ratio (PSNR) and Structural Similarity Index (SSIM) are adopted for quantitative evaluation. We also compare our method on real-world hazy images to test the generalization ability of our model \cite{92}.\\

\begin{figure}[h]
	\centering
	\includegraphics[width=1\columnwidth]{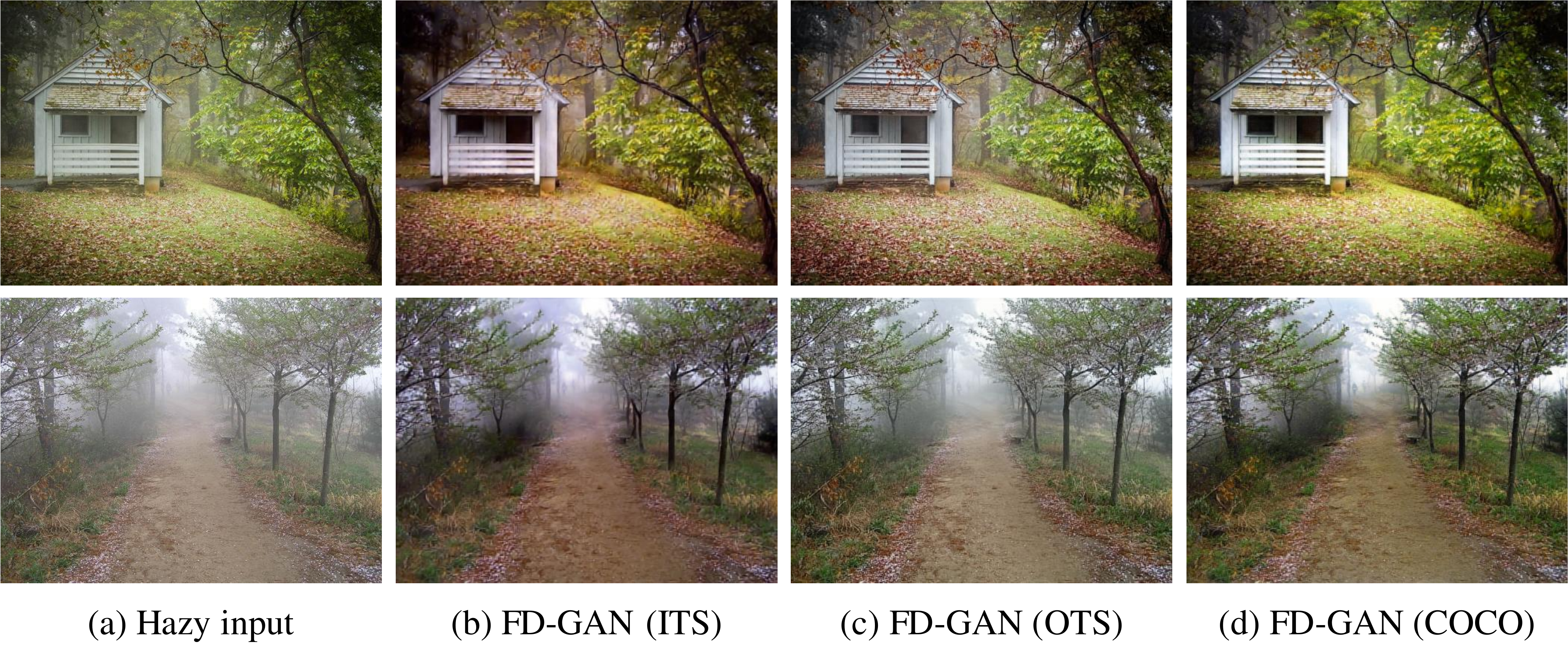}	
	\centering
	\caption{Visual comparsions on different training datasets.}
	\label{figure2}
\end{figure}

\subsection{Implementation Details}
In our experiments, the loss weights  $\alpha_1$, $\alpha_2$, $\alpha_3$ and $\alpha_4$ are set to 2, 1, 2,  and 0.1. For training, the input size is $256 \times 320 \times$ 3. To extract LF, the window size of Gaussian filter is set to 15 with standard deviation $\sigma=3$. As for HF, for convenience and stability, we first transform the RGB image into gray image then apply Laplacian operator on it, with the kernel size of 3. In training processing, we use Adam optimizer with initial learning rate of $2 \times 10^{-3}$ for both generator and discriminator. The network was trained for totally 435,600 iterations by Pytorch with an Nvidia RTX 2080Ti GPU. The inference time can reach 65 FPS with an RGB input size of $1024 \times 1024$.

\subsection{Comparison with state-of-the-art methods}
We compare the proposed method with several state-of-the-art methods: DCP \cite{3}, AOD \cite{40}, DCPDN \cite{16}, GFN \cite{42} and EPDN \cite{75}. The evaluation is conducted on both synthetic datasets and real-world images. We run the released codes of these methods and adopt the evaluation toolkit from NTIRE'18 Challenge.\\
\textbf{Performance on synthetic datasets.} We use Peak Signal to Noise Ratio (PSNR) and Structural Similarity Index (SSIM) for quantitative evaluation. The results are shown in Table 2. It can be seen that our method reaches the highest values in both PSNR and SSIM by a large margin. As shown in Figure 4, DCP \cite{3} fails in the sky area. The results of AOD \cite{40} and DCPDN \cite{16} leave haze residuals in some areas. The methods of GFN \cite{42} and EPDN \cite{75} remove the haze well but bring about exorbitant contrast. In contrast, our method can generate more natural results with sharper textures and better color fidelity, which are shown to be more visually faithful to the ground-truth.\\
\textbf{Performance on real-world datasets.} We also evaluate the results on real-world hazy images and observe that although the model is trained with synthetic images, it can generalize well on natural hazy images. In Figure 5, we choose several commonly used hazy images for comparison. It can be observed that AOD and DCPDN leave lots of haze residuals and cause slight color distortion. DCP leaves a little haze in results and tends to darken the images. GFN and EPDN are unable to remove haze thoroughly in the heavily hazy area. Comparing with those methods, our model can not only remove haze much more thoroughly, whatever in the slight or heavy haze area, but also generate much sharper details with less color distortion and few artifacts.

\section{Ablation Study}
\textbf{Effect of training datasets.} To reveal the influence of training data, we investigate the performance of models trained on different datasets. We train our model with three different datasets: 1) our synthetic dataset. 2) Indoor Training Set (ITS) of RESIDE dataset. 3) Outdoor Training Set (OTS) of RESIDE dataset, which is synthesized from NYU. The comparison results are displayed in Figure 7 and the last three rows of Table 2. It shows that the model trained on our synthetic dataset surpasses models trained on other datasets qualitatively and quantitively.\\
\textbf{Effect of the proposed Fusion-discriminator.} To demonstrate the effectiveness and superiority of the proposed Fusion-discriminator, we conduct an ablation study by gradually modifying the baseline model and comparing their performance. Specifically, there are three different designs: 1) Baseline model with only L1 loss and SSIM loss, 2) Introduce standard GAN structure, 3) The proposed method with the Fusion-discriminator (Fusion-full model). To make fair comparison, we keep the same network architecture and training settings for all models above, except the modification depicted in Table 3.

\linespread{1.12}
\begin{table}[t]
	\setlength{\abovecaptionskip}{0.235cm}
	\setlength{\belowcaptionskip}{-0cm}
	\newcommand{\tabincell}[2]{\begin{tabular}{@{}#1@{}}#2\end{tabular}}  
	\centering
	\caption{Ablation study results. By developing the Fusion-discriminator, our method obtains significant improvements.}\smallskip	
	\begin{tabular}{p{48pt}<{\centering}p{9pt}|p{31pt}<{\centering}p{31pt}<{\centering}p{45pt}<{\centering}}
		\hline
		& & Baseline & GAN & Ours\\ \hline 
		\multicolumn{2}{c|}{L1 loss + SSIM loss} &  ${\surd}$ & ${\surd}$ & ${\surd}$ \\
		\multicolumn{2}{c|}{Adversarial loss} & ${\times}$ & ${\surd}$ & ${\surd}$ \\
		\multicolumn{2}{c|}{Fusion-discriminator} & ${\times}$ & ${\times}$ & ${\surd}$ \\ \hline \hline
		\multirow{2}{*}{\tabincell{c}{SOTS}}  & \underline{PSNR} & 18.5145 & 21.6263 & \textbf{23.1529} \\ 
		& \underline{SSIM} & 0.8714 & 0.8922 & \textbf{0.9207} \\ \hline
		\multirow{2}{*}{NTIRE'18} & \underline{PSNR} &  16.8574 & 16.5912 & \textbf{18.0684}\\
		& \underline{SSIM} &  0.6952 & 0.7116 & \textbf{0.7300} \\
		\hline
	\end{tabular}
	\label{tab:my_label}
\end{table}

\begin{figure}[h]
	\centering
	\includegraphics[width=1\columnwidth]{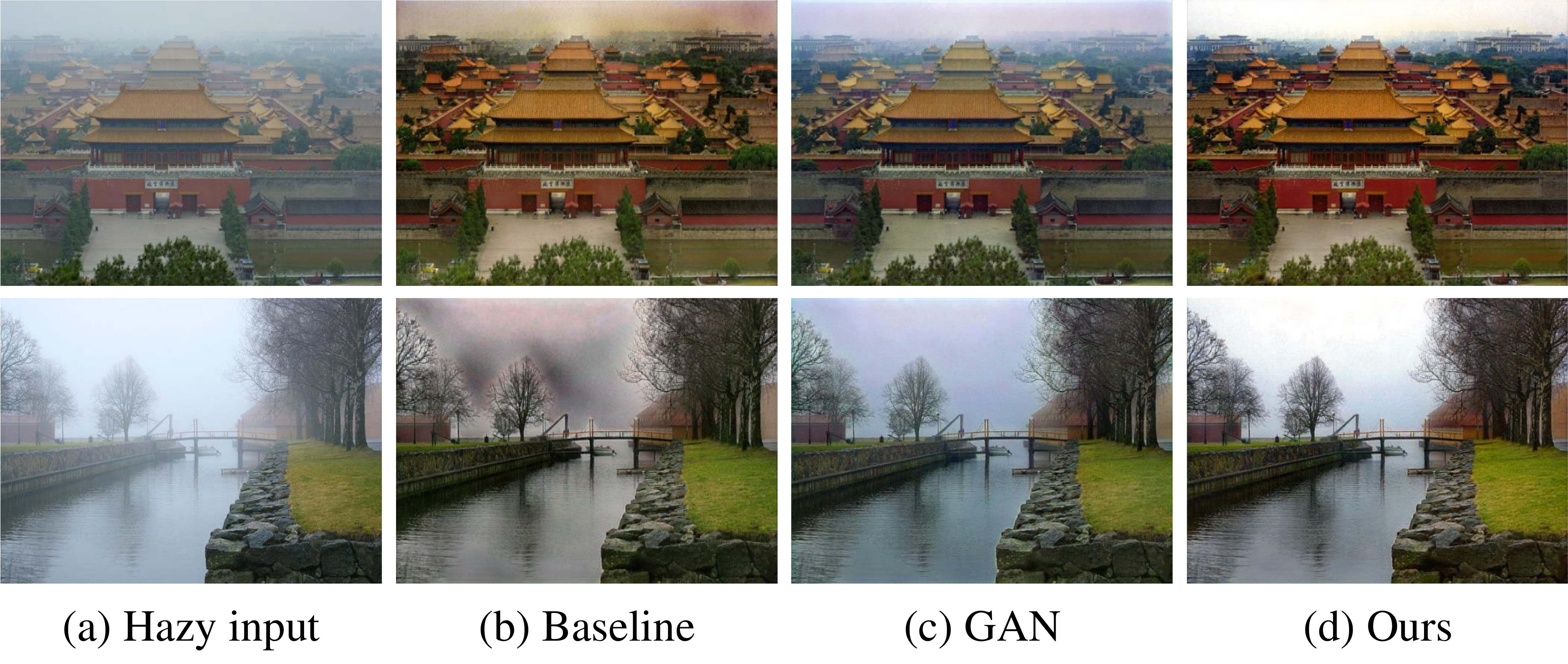}	
	
	\caption{By introducing the proposed Fusion-discriminator, our model can generate more natural dehazed images.}
	\label{figure2}
\end{figure}

We evaluate those models on SOTS and NTIRE'18 Test sets. The results are displayed in Table 3. It is shown that models with GAN surpass baseline which only adopts L1 loss and SSIM loss in terms of PSNR/SSIM. Further, as shown in Figure 8, the dehazed results of baseline model contain lots of haze residuals while standard GAN model is able to remove more haze effect but tends to produce some artifacts and color distortion. By introducing the Fusion-discriminator, our model can remove haze more thoroughly and generate cleaner images with better color fidelity.

\section{Conclusion}
In this paper, we propose a fully end-to-end algorithm FD-GAN for image dehazing. Moreover, we develop a novel Fusion-discriminator which can integrate the frequency information as additional priors and constraints into the dehazing network. Our method can generate more visually pleasing dehazed results with less color distortion. Extensive experimental results have demonstrated that our method performs favorably against several state-of-the-art methods on both synthetic datasets and real-world hazy images.

\section{Acknowledgement}
This work is supported by National Natural Science Foundation of China (U1713203) and Shenzhen Science and Technology Innovation Commission (Project KQJSCX20180330170238897).

\bibliographystyle{aaai} 
\bibliography{bibliography}
\end{document}